\pdfoutput=1

\documentclass[11pt,table,dvipsnames]{article}

\usepackage{ACL2023}

\usepackage{times}
\usepackage{latexsym}

\usepackage{tabularx}
\usepackage{booktabs}
\usepackage{fontawesome5}
\usepackage{graphicx}
\usepackage{multirow}
\usepackage{tikz-dependency}
\usepackage{subcaption}
\usepackage{tabularx}
\usepackage{booktabs}
\usepackage{caption}
\usepackage{multirow}
\usepackage{makecell}
\usepackage{nicefrac}
\usepackage{xfrac}
\usepackage{amssymb}
\usepackage[flushleft]{threeparttable}
\usepackage{soul}
\usepackage{tablefootnote}
\usepackage{xcolor}
\usepackage{tabularx}
\newcolumntype{Y}{>{\centering\arraybackslash}X}

\newcommand*\circled[1]{\tikz[baseline=(char.base)]{
            \node[shape=circle,draw,inner sep=.6pt] (char) {#1};}}

\newcommand{\green}[1]{\cellcolor{ForestGreen!#1}}
\newcommand{\yellow}[1]{\cellcolor{Goldenrod!#1}}

\usepackage[T1]{fontenc}

\usepackage[utf8]{inputenc}

\usepackage{microtype}

\usepackage{inconsolata}

\title{Silver Syntax Pre-training for Cross-Domain Relation Extraction}

\author{Elisa Bassignana\textsuperscript{\faCompass} \hspace{.8em}
Filip Ginter\textsuperscript{\faCommentDots[regular]} \hspace{.8em}
Sampo Pyysalo\textsuperscript{\faCommentDots[regular]} \\
{\bf Rob van der Goot}\textsuperscript{\faCompass} \hspace{.8em}
{\bf Barbara Plank}\textsuperscript{\faCompass}\textsuperscript{\faMountain}\\
\textsuperscript{\faCompass}Department of Computer Science, IT University of Copenhagen, Denmark \\ 
\textsuperscript{\faCommentDots[regular]}TurkuNLP, Department of Computing, University of Turku, Finland \\
\textsuperscript{\faMountain}MaiNLP, Center for Information and Language Processing, LMU Munich, Germany \\
\texttt{elba@itu.dk}}

\begin{document}
\maketitle
\begin{abstract}

Relation Extraction (RE) remains a challenging task, especially when considering realistic out-of-domain evaluations.
One of the main reasons for this is the limited training size of current RE datasets: obtaining high-quality (manually annotated) data is extremely expensive and cannot realistically be repeated for each new domain. 
An intermediate training step on data from related tasks has shown to be beneficial across many NLP tasks.
However, this setup still requires supplementary annotated data, which is often not available.
In this paper, we investigate intermediate pre-training specifically for RE. We exploit the affinity between syntactic structure and semantic RE, and identify the syntactic relations which are closely related to RE by being on the shortest dependency path between two entities. 
We then take advantage of the high accuracy of current syntactic parsers in order to automatically obtain large amounts of low-cost pre-training data. 
By pre-training our RE model on the relevant syntactic relations, we are able to outperform the baseline in five out of six cross-domain setups, \textit{without} any additional annotated data.
 
\end{abstract}

\section{Introduction}

Relation Extraction (RE) is the task of extracting structured knowledge, often in the form of triplets, from unstructured text.
Despite the increasing attention this task received in recent years, the performance obtained so far are very low~\cite{popovic-farber-2022-shot}. This happens in particular when considering realistic scenarios which include out-of-domain setups, and deal with the whole task---in contrast to the simplified Relation Classification which assumes that the correct entity pairs are given~\cite{han-etal-2018-fewrel,baldini-soares-etal-2019-matching,gao-etal-2019-fewrel}.
One main challenge of RE and other related Information Extraction tasks is the "domain-specificity": Depending on the text domain, the type of information to extract changes.
For example, while in the news domain we can find entities like \textit{person} and \textit{city}, and relations like \textit{city of birth}~\cite{zhang-etal-2017-position}, in scientific texts, we can find information about \textit{metrics}, \textit{tasks} and \textit{comparisons} between computational models~\cite{luan-etal-2018-multi}.
While high-quality, domain-specific data for fine-tuning the RE models would be ideal, as for many other NLP tasks, annotating data is expensive and time-consuming.\footnote{For example,~\citealp{bassignana-plank-2022-crossre} report a cost of 19K USD ( $\approx$ 1\$ per annotated relation) and seven months of annotation work for an RE dataset including 5.3K sentences.}
A recent approach that leads to improved performance on a variety of NLP tasks is intermediate task training. It consists of a step of training on one or more NLP tasks between the general language model pre-training and the specific end task fine-tuning (STILT, Supplementary Training on Intermediate Labeled-data Tasks;~\citealp{stilt}).
However, STILT assumes the availability of additional high quality training data, annotated for a related task.

\begin{figure}
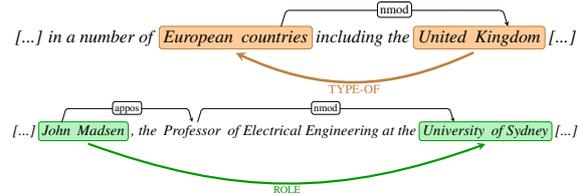

\centering
\resizebox{\columnwidth}{!}{
\begin{dependency}
\begin{deptext}
\textit{[...] in a number of} \& \textit{European} \& \textit{countries} \& \textit{including the} \& \textit{United} \& \textit{Kingdom} \& \textit{[...]} \\ \\
\end{deptext}
\wordgroup[group style={fill=orange!40, draw=brown}]{1}{2}{3}{entity2}
\wordgroup[group style={fill=orange!40, draw=brown}]{1}{5}{6}{entity1}
\depedge[edge height=2ex]{3}{6}{nmod}
\groupedge[edge below, edge style = {ultra thick, brown}, arc edge, arc angle=25, label style = {below, text=brown}, text only label]{entity1}{entity2}{TYPE-OF}{1ex}
\end{dependency}}\\

\resizebox{\columnwidth}{!}{
\begin{dependency}
\begin{deptext}
\textit{[...]} \& \textit{John} \& \textit{Madsen} \& \textit{, the} \& \textit{Professor} \& \textit{of Electrical Engineering at the} \& \textit{University} \& \textit{of Sydney} \& \textit{[...]} \\ \\
\end{deptext}
\wordgroup[group style={fill=green!85!blue!30!white, draw=green!60!black}]{1}{2}{3}{entity1}
\wordgroup[group style={fill=green!85!blue!30!white, draw=green!60!black}]{1}{7}{8}{entity2}
\depedge[edge height=2ex]{2}{5}{appos}
\depedge[edge height=2ex]{5}{7}{nmod}
\groupedge[edge below, edge style = {ultra thick, green!60!black}, arc edge, arc angle=20, label style = {below, text=green!60!black}, text only label]{entity1}{entity2}{ROLE}{1ex}
\end{dependency}}

\caption{\label{fig:fig1}\textbf{Syntactic and Semantic Structures Affinity.} Shortest dependency path (above), and semantic relation (below) between two semantic entities.}
\end{figure}

\begin{figure*}
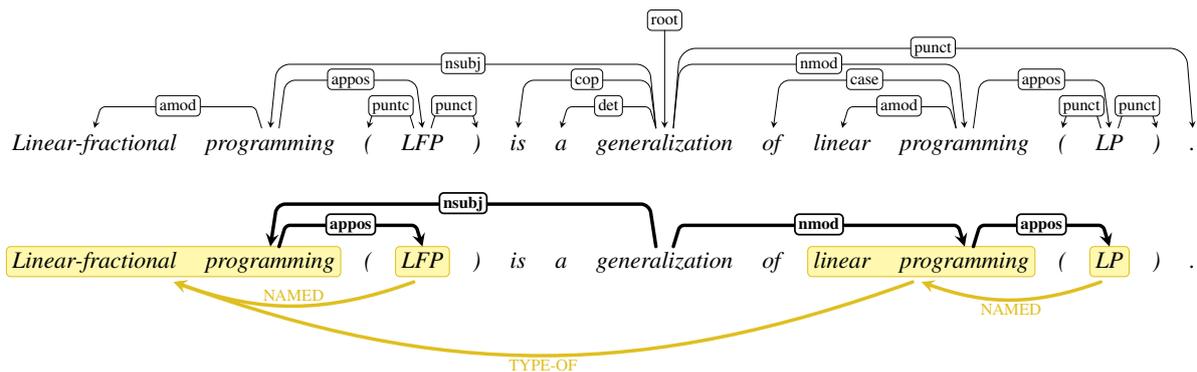

\centering
\resizebox{\textwidth}{!}{
\begin{dependency}
\begin{deptext}[column sep=0.3cm]
\textit{Linear-fractional} \& \textit{programming} \& \textit{(} \& \textit{LFP} \& \textit{)} \& \textit{is} \& \textit{a} \& \textit{generalization} \& \textit{of} \& \textit{linear} \& \textit{programming} \& \textit{(} \& \textit{LP} \& \textit{)} \& \textit{.} \\ \\
\end{deptext}
\depedge[edge height=2ex]{2}{1}{amod}
\depedge[edge height=6ex]{8}{2}{nsubj}
\depedge[edge height=2ex]{4}{3}{puntc}
\depedge[edge height=4.5ex]{2}{4}{appos}
\depedge[edge height=2ex]{4}{5}{punct}
\depedge[edge height=4.5ex]{8}{6}{cop}
\depedge[edge height=2ex]{8}{7}{det}
\deproot{8}{root}
\depedge[edge height=4.5ex]{11}{9}{case}
\depedge[edge height=2ex]{11}{10}{amod}
\depedge[edge height=6ex]{8}{11}{nmod}
\depedge[edge height=2ex]{13}{12}{punct}
\depedge[edge height=4.5ex]{11}{13}{appos}
\depedge[edge height=2ex]{13}{14}{punct}
\depedge[edge height=7.5ex]{8}{15}{punct}
\end{dependency}} \\

\resizebox{\textwidth}{!}{
\begin{dependency}
\begin{deptext}[column sep=0.3cm]
\textit{Linear-fractional} \& \textit{programming} \& \textit{(} \& \textit{LFP} \& \textit{)} \& \textit{is} \& \textit{a} \& \textit{generalization} \& \textit{of} \& \textit{linear} \& \textit{programming} \& \textit{(} \& \textit{LP} \& \textit{)} \& \textit{.} \\ \\
\end{deptext}
\depedge[edge height=4ex, edge style={ultra thick}, label style={font=\bfseries,thick}]{8}{2}{nsubj}
\depedge[edge height=2ex, edge style={ultra thick}, label style={font=\bfseries,thick}]{2}{4}{appos}
\depedge[edge height=2ex, edge style={ultra thick}, label style={font=\bfseries,thick}]{8}{11}{nmod}
\depedge[edge height=2ex, edge style={ultra thick}, label style={font=\bfseries,thick}]{11}{13}{appos}
\wordgroup[group style={fill=yellow!40, draw=brown!50!yellow}]{1}{1}{2}{entity1}
\wordgroup[group style={fill=yellow!40, draw=brown!50!yellow}]{1}{4}{4}{entity2}
\wordgroup[group style={fill=yellow!40, draw=brown!50!yellow}]{1}{10}{11}{entity3}
\wordgroup[group style={fill=yellow!40, draw=brown!50!yellow}]{1}{13}{13}{entity4}
\groupedge[edge below, edge style = {ultra thick, brown!50!yellow}, arc edge, arc angle=20, label style = {above, text=brown!50!yellow}, text only label]{entity2}{entity1}{NAMED}{1ex}
\groupedge[edge below, edge style = {ultra thick, brown!50!yellow}, arc edge, arc angle=20, label style = {below, text=brown!50!yellow}, text only label]{entity3}{entity1}{TYPE-OF}{1ex}
\groupedge[edge below, edge style = {ultra thick, brown!50!yellow}, arc edge, arc angle=20, label style = {below, text=brown!50!yellow}, text only label]{entity4}{entity3}{NAMED}{1ex}
\end{dependency}}
\caption{\label{fig:pre-train}\textbf{Pre-training Example}. Given the dependency tree (above), we filter in for pre-training only the UD labels which are on the shortest dependency path between two semantic entities (below).}
\end{figure*}

In this paper, we explore intermediate pre-training specifically for cross-domain RE and look for alternatives which avoid the need of external manually annotated datasets to pre-train the model on.
In particular, we analyze the affinity between syntactic structure and semantic relations, by considering the shortest dependency path between two entities~\cite{bunescu-mooney-2005-shortest,10.1093/bioinformatics/btl616,bjorne-etal-2009-extracting,liu-etal-2015-dependency}.
We replace the traditional intermediate pre-training step on additional annotated data, with a \textit{syntax pre-training} step on silver data.
We exploit the high accuracy of current syntax parsers, for obtaining large amount of low-cost pre-training data. 
The use of syntax has a long tradition in RE~\cite{zhang-etal-2006-exploring,qian-etal-2008-exploiting, nguyen-etal-2009-convolution, peng-etal-2015-extended}. Recently, work has started to infuse syntax during language model pre-training~\cite{sachan-etal-2021-syntax} showing benefits for RE as well. We instead investigate dependency information as silver data in intermediate training, which is more efficient. To the best of our knowledge, the use of syntax in intermediate pre-training for RE is novel.
We aim to answer the following research questions: \circled{1} Does syntax help RE via intermediate pre-training (fast and cheap approach)? and \circled{2} How does it compare with pre-training on additional labeled RE data (expensive)? We release our model and experiments.\footnote{\url{https://github.com/mainlp/syntax-pre-training-for-RE}}

\section{Syntax Pre-training for RE}
\label{sec:idea-explained}

Syntactic parsing is a structured prediction task aiming to extract the syntactic structure of text, most commonly in the form of a tree.
RE is also a structured prediction task, but with the aim of extracting the semantics expressed in a text in the form of triplets---entity A, entity B, and the semantic relation between them.\footnote{In this project, we follow previous work, and assume gold entities, leaving end-to-end RE for future work.}
We exploit the affinity of these two structures by considering the shortest dependency path between two (semantic) entities (see Figure~\ref{fig:fig1}).

The idea we follow in this work is to pre-train an RE baseline model over the syntactic relations---Universal Dependency (UD) labels---which most frequently appear on the shortest dependency paths between two entities (black bold arrows in Figure~\ref{fig:pre-train}).
We assume these labels to be the most relevant with respect to the final target task of RE.
In order to feed the individual UD relations into the RE baseline (model details in Section~\ref{sec:exp-setup}) we treat them similarly as the semantic connections. 
In respect to Figure~\ref{fig:pre-train}, we can formalize the semantic relations as the following triplets:

\begin{itemize}
\setlength\itemsep{0em}
\small
    \item \texttt{NAMED(LFP,Linear-fractional programming)}
    \item \texttt{TYPE-OF(linear programming,Linear-fractional programming)}
    \item \texttt{NAMED(LP,linear programming)}.
\end{itemize}
Accordingly, we define the syntax pre-training instances as:
\begin{itemize}
\setlength\itemsep{0em}
\small
    \item \texttt{appos(programming,LFP)}
    \item \texttt{nsubj(generalization,programming)}
    \item \texttt{nmod(generalization,programming)}
    \item \texttt{appos(programming,LP)}.
\end{itemize}

In the next section we describe the detailed training process.

\section{Experiments}

\subsection{Setup}
\label{sec:exp-setup}

\begin{figure*}
    \centering
    \includegraphics[width=\textwidth]{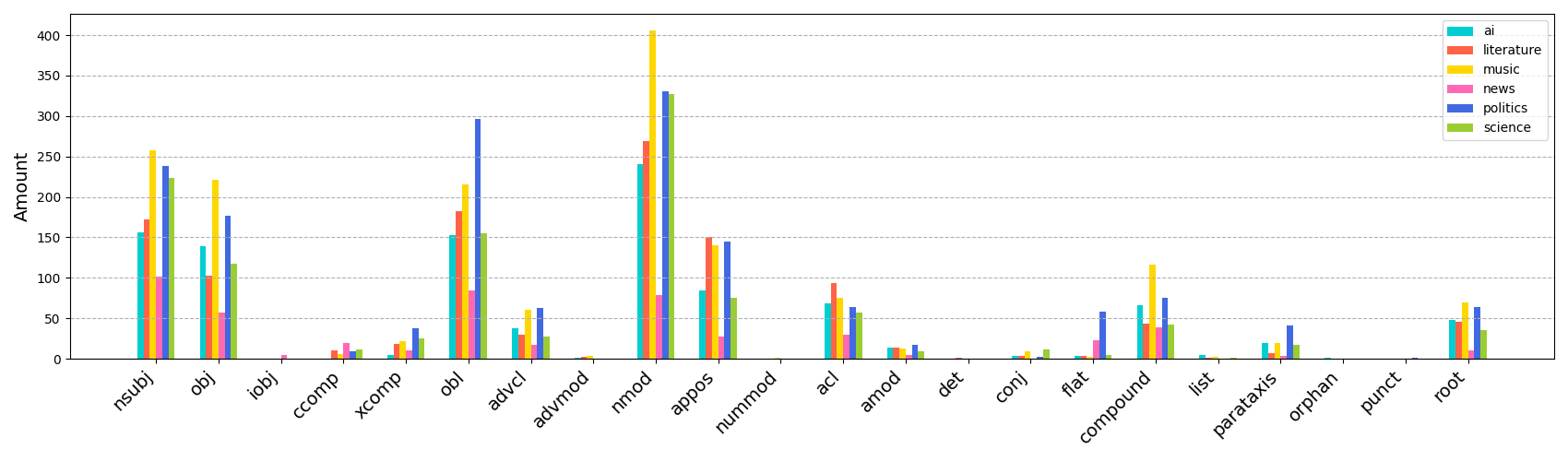}
    \caption{\textbf{UD Label Distribution Over the Shortest Dependency Paths.} Statistics of the UD labels which are on the shortest dependency path between two entities over the six train sets of CrossRE~\cite{bassignana-plank-2022-crossre}.}
    \label{fig:ud-labels}
\end{figure*}

\begin{figure}
    \centering
    \includegraphics[width=\columnwidth]{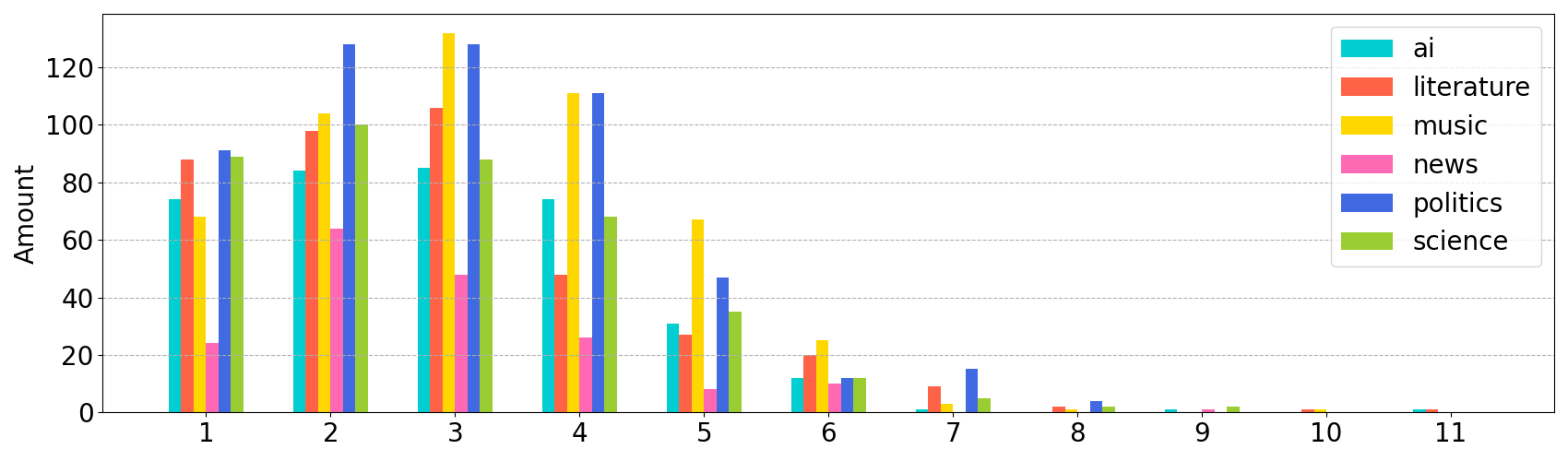}
    \caption{\textbf{Shortest Dependency Path Length.} Statistics of the shortest dependency path length between two entities over the train sets of CrossRE~\cite{bassignana-plank-2022-crossre}.}
    \label{fig:path-length}
\end{figure}

\paragraph{Data}
In order to evaluate the robustness of our method over out-of-domain distributions, we experiment with CrossRE~\cite{bassignana-plank-2022-crossre},\footnote{Released with a GNU General Public License v3.0.} a recently published multi-domain dataset.
CrossRE includes 17 relation types spanning over six diverse text domains: news, politics, natural science, music, literature and artificial intelligence (AI). The dataset was annotated on top of a Named Entity Recognition dataset---CrossNER~\cite{crossNER}---which comes with an unlabeled domain-related corpora.\footnote{Released with an MIT License.}
We used the latter for the \textit{syntax pre-training} phase.

\paragraph{UD Label Selection}
In order to select the UD labels which most frequently appear on the shortest dependency path between two semantic entities, we parsed the training portions of CrossRE.
Our analysis combines RE annotations and syntactically parsed data.
We observe that the syntactic distance between two entities is often higher than one (see Figure~\ref{fig:path-length}), meaning that the shortest dependency path between two entities includes multiple dependencies---in the examples in Figure~\ref{fig:fig1}, the one above has distance one, the one below has distance two.
However, the shortest dependency paths contain an high frequency of just a few UD labels (see Figure~\ref{fig:ud-labels}) which we use for \textit{syntax pre-training}: \texttt{nsubj}, \texttt{obj}, \texttt{obl}, \texttt{nmod}, \texttt{appos}.
See Appendix~\ref{app:ud-analysis} for additional data analysis.

\paragraph{Model}
Our RE model follows the current state-of-the-art architecture by \citealp{baldini-soares-etal-2019-matching} which augments the sentence with four entity markers $e_1^{start}$, $e_1^{end}$, $e_2^{start}$, $e_2^{end}$ before feeding it into a pre-trained encoder (BERT;~\citealp{devlin-etal-2019-bert}).
The classification is then made by a 1-layer feed-forward neural network over the concatenation of the start markers $[\hat{s}_{e_1^{start}}, \hat{s}_{e_2^{start}}]$.
We run our experiments over five random seeds and report the average performance. See Appendix~\ref{app:reproducibility} for reproducibility and hyperparameters settings of our model.

\paragraph{Training}
The training of our RE model is divided into two phases.
In the first one---which we are going to call \textit{syntax pre-training}---we use the unlabeled corpora from CrossNER for pre-training the baseline model over the \textit{RE-relevant} UD labels.
To do so, \circled{1} we sample an equal amount of sentences from each domain\footnote{Regarding the news domain, which does not have a corresponding unlabeled corpus available, we sampled from the train set of CrossNER which is not included in CrossRE.} (details in Section~\ref{sec:analysis}), and \circled{2} use the MaChAmp toolkit~\cite{van-der-goot-etal-2021-massive} for inferring the syntactic tree of each of them.
We apply an additional sub-step for disentangling the \texttt{conj} dependency, as illustrated in Appendix~\ref{app:conj}.
Then, \circled{3} we filer in only the \texttt{nsubj}, \texttt{obj}, \texttt{obl}, \texttt{nmod}, and \texttt{appos} UD labels and \circled{4} feed those connections to the RE model (as explained in the previous section).
Within the RE model architecture described above, each triplet corresponds to one instance. In this phase, in order to assure more variety, we randomly select a maximum of five triplets from each pre-train sentence.

In the second training phase---the \textit{fine-tuning} one---we replace the classification head (i.e.\ the feed-forward layer) with a new one, and individually train six copies of the model over the six train sets of CrossRE. Note that the encoder is fine-tuned in both training phases.
Finally, we test each model on in- and out-of-domain setups. 

\subsection{Results}

Table~\ref{tab:results} reports the results of our cross-domain experiments in terms of Macro-F1.
We compare our proposed approach which adopts \textit{syntax pre-training} with the zero-shot baseline model.\footnote{While utilizing the model implementation by~\citealp{bassignana-plank-2022-crossre}, our score range is lower because we include the \textit{no-relation} case, while they assume gold entity pairs.} Five out of six models outperform the average of the baseline evaluation, including in- and out-of-domain assessments. The average improvement---obtained without any additional annotated RE data---is 0.71, which considering the low score range given by the challenging dataset (with limited train sets, see dataset size in Appendix~\ref{app:crossre}), and the cross-domain setup, is considerable.
The model fine-tuned on the news domain is the only one not outperforming the baseline. However, the performance scores on this domain are already extremely low for the baseline, because news comes from a different data source with respect to the other domains, has a considerable smaller train set, and present a sparse relation types distribution, making it a bad candidate for transferring to other domains~\cite{bassignana-plank-2022-crossre}.

\begin{table}[]
    \centering
    \resizebox{\columnwidth}{!}{
    \begin{tabular}{c|r|rrrrrr|r}
         \toprule
          & & \multicolumn{7}{c}{\textsc{test}} \\
          & \textsc{train} & \multicolumn{1}{c}{news} & \multicolumn{1}{c}{politics} & \multicolumn{1}{c}{science} & \multicolumn{1}{c}{music} & \multicolumn{1}{c}{literature} & \multicolumn{1}{c}{AI} & \multicolumn{1}{c}{avg.}\\
         \midrule
         \multirow{6}{*}{\rotatebox{90}{\textsc{baseline}}} & news & 10.98 & 1.32 & 1.24 & 1.01 & 1.49 & 1.42 & \textbf{2.91} \\
         & politics & 16.07 & 11.30 & 6.74 & 7.24 & 7.29 & 5.54 & 9.03 \\
         & science & 6.54 & 5.95 & 8.57 & 7.13 & 6.65 & 7.29 & 7.02 \\
         & music & 3.99 & 9.91 & 9.22 & 19.01 & 10.43 & 8.53 & 10.18 \\
         & literature & 11.30 & 9.60 & 9.79 & 12.49 & 17.17 & 9.79 & 11.69 \\
         & AI & 6.58 & 7.42 & 11.03 & 7.11 & 6.15 & 15.57 & 8.98 \\
         \midrule
         \multirow{6}{*}{\rotatebox{90}{\textsc{syntax}}} & news & 6.67 & 1.15 & 0.72 & 0.61 & 1.13 & 0.75 & 1.84 \\
         & politics & 13.72 & 12.09 & 7.47 & 7.15 & 7.78 & 6.24 & \textbf{9.08} \\
         & science & 8.46 & 7.08 & 8.69 & 8.19 & 7.52 & 8.91 & \textbf{8.14} \\
         & music & 3.35 & 10.65 & 9.35 & 18.63 & 11.62 & 10.34 & \textbf{10.66} \\
         & literature & 11.85 & 9.84 & 10.35 & 13.58 & 18.64 & 9.94 & \textbf{12.37} \\
         & AI & 8.87 & 8.59 & 11.87 & 8.29 & 7.68 & 15.93 & \textbf{10.21} \\
         \midrule
         \midrule
         \multirow{6}{*}{\rotatebox{90}{\textsc{scierc}}} & news & 11.88 & 2.30 & 2.09 & 1.13 & 1.82 & 2.16 & 3.56 \\
         & politics & 14.25 & 13.55 & 6.52 & 7.12 & 7.42 & 7.07 & 9.32 \\
         & science & 8.27 & 10.31 & 13.59 & 9.09 & 7.78 & 11.11 & 10.03 \\
         & music & 5.41 & 11.84 & 10.85 & 21.39 & 12.26 & 11.22 & 12.16 \\
         & literature & 12.36 & 8.05 & 8.87 & 13.13 & 16.44 & 9.40 & 11.37 \\
         & AI & 11.00 & 10.12 & 14.03 & 8.93 & 8.50 & 18.89 & 11.91 \\
         \bottomrule
    \end{tabular}}
    \caption{\textbf{Performance Scores.} Macro-F1 scores of the baseline model, compared with the proposed \textit{syntax pre-training} approach, and---as comparison---with the traditional pre-training over the manually annotated SciERC dataset~\cite{luan-etal-2018-multi}.}
    \label{tab:results}
\end{table}

As comparison, we report the scores obtained with the traditional intermediate pre-training which includes additional annotated data. We pre-train the language encoder on SciERC~\cite{luan-etal-2018-multi}, a manually annotated dataset for RE. 
SciERC contains seven relation types, of which three overlap with the CrossRE relation set. In this setup, the improvement over the baseline includes the news, but not the literature domain. Nevertheless, while the gain is on average slightly higher with respect to the proposed \textit{syntax pre-training} approach, it comes at a much higher annotation cost. 

\begin{figure}
    \centering
    \includegraphics[width=\columnwidth]{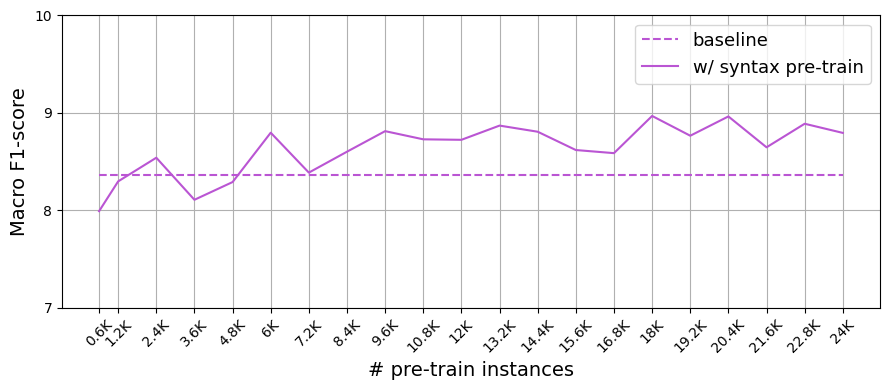}
    \caption{\textbf{Pre-train Data Quantity Analysis.} Average (dev) performance of the six models when pre-trained over increasing amounts of syntactic instances.}
    \label{fig:pretrain}
\end{figure}

\begin{figure}
\centering
\includegraphics[width=\columnwidth]{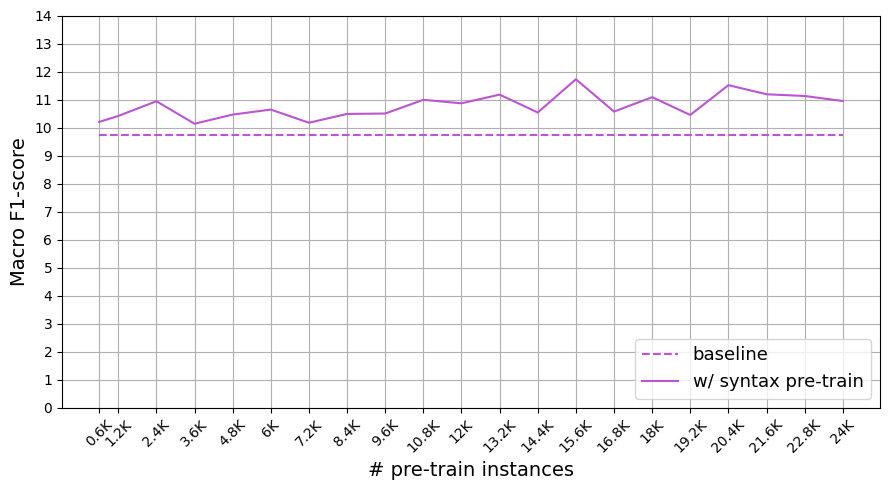}
\caption{\textbf{Per-Domain Pre-train Data Quantity Analysis.} Individual (dev) performance of the model fine-tuned on AI when pre-trained over increasing amounts of syntactic instances.}
\label{fig:ai-per-domain}
\end{figure}

\section{Pre-training Data Quantity Analysis}
\label{sec:analysis}

We inspect the optimal quantity of syntactic data to pre-train our RE model on by fine-tuning this hyperparameter over the dev sets of CrossRE. The plot in Figure~\ref{fig:pretrain} reports the average performance of the six models when pre-trained on increasing amounts of syntactic dependencies.\footnote{Pre-training performance in Appendix~\ref{app:paser}.} Starting from 8.4K instances onward, the performance stabilizes above the baseline. We select the peak (20.4K, albeit results are similar between 18-20.4K) for reporting our test set results in Table~\ref{tab:results}.
While we are interested in the robustness of our method across multiple domains, and therefore consider the average (Figure~\ref{fig:pretrain}), domain-optima could be achieved by examining individual domain performance.
As example, we report in Figure~\ref{fig:ai-per-domain} the plot relative to the model fine-tuned on AI, which is the one  obtaining the highest gain.
The model fine-tuned on AI generally gains a lot from the \textit{syntax pre-training} step, with its peak on 15.6K pre-training instances.

\section{Conclusion}

We introduce \textit{syntax pre-training} for RE as an alternative to the traditional intermediate training which uses additional manually annotated data.
We pre-train our RE model over silver UD labels which most frequently connect the semantic entities via the shortest dependency path.
We test the proposed method over CrossRE and outperform the baseline in five out of six cross-domain setups.
Pre-training over a manually annotated dataset, in comparison, only slightly increases our scores in five out of six evaluations, but at a much higher cost.

\section*{Limitations}

While we already manage to outperform the baseline, the pre-training data quantity is relatively small ($\sim$20K instances).
Given the computational cost of training 30 models---six train sets, over five random seeds each---and testing them within in- and cross- domain setups, we break the inspection of the optimal pre-training data amount at 24K instances.
However we do not exclude that more pre-training instances would be even more beneficial for improving even more over the baseline.

Related to computation cost constrains, we test our \textit{syntax pre-training} approach over one set of UD labels only (\texttt{nsubj}, \texttt{obj}, \texttt{obl}, \texttt{nmod}, \texttt{appos}). Different sets could be investigated, e.g.\ including \texttt{acl} and \texttt{compound}, which present a lower, but still considerable amount of instances (see Figure~\ref{fig:ud-labels}).

Finally, while approaching RE by assuming that the gold entities are given is a common area of research, we leave for future work the inspection of the proposed method over end-to-end RE.

\section*{Acknowledgments}
We thank the NLPnorth and the MaiNLP groups for feedback on an earlier version of this paper, and TurkuNLP for hosting EB for a research stay.

EB and BP are supported by the Independent Research Fund Denmark (Danmarks Frie Forskningsfond; DFF) Sapere Aude grant 9063-00077B. BP is in parts supported by the European Research Council (ERC)  (grant agreement No.\ 101043235). FG and SP were supported by the Academy of Finland.

\bibliography{anthology,custom}
\bibliographystyle{acl_natbib}

\appendix

\section{UD Analysis for RE}
\label{app:ud-analysis}

We inspect the same statistics as Figure~\ref{fig:ud-labels} and Figure~\ref{fig:path-length}---UD labels on the shortest dependency paths, and shortest dependency path lengths respectively---but instead of at domain level, at semantic relation type level. 
Table~\ref{fig:per-label-ud} and Table~\ref{fig:per-label-sdp} report this analysis, revealing similar trends over the 17 types.

\begin{table*}
\centering
\setlength{\tabcolsep}{0.1em}
\tiny
\begin{tabularx}{\textwidth}{|c|r|YYYYYYYYYYYYYYYYY|}
\toprule
\multicolumn{2}{c|}{} & rel-to & artifact & cause-eff & compare & gen-aff & named & opposite & origin & part-of & physical & role & social & temporal & topic & type-of & usage & win-def \\
\midrule

\multirow{21}{*}{\rotatebox{90}{\textbf{\fontsize{5}{5}\selectfont Universal Dependencies}}}
& nsubj & \yellow{90}89 & \yellow{80}106 & \yellow{50}2 & \yellow{90}12 & \yellow{90}120 & \yellow{65}54 & \yellow{100}61 & \yellow{90}53 & \yellow{70}75 & \yellow{50}115 & \yellow{90}248 & \yellow{100}33 & \yellow{50}54 & \yellow{30}10 & \yellow{50}18 & \yellow{90}30 & \yellow{70}68 \\
& obj   & \yellow{70}78 & \yellow{30}51 & \yellow{00}1 & \yellow{50}6 & \yellow{50}76 & \yellow{40}36 & \yellow{80}48 & \yellow{60}41 & \yellow{80}83 & \yellow{10}55 & \yellow{50}129 & \yellow{20}9 & \yellow{40}48 & \yellow{50}17 & \yellow{30}14 & \yellow{100}37 & \yellow{100}86 \\
& iobj  & \yellow{00}0 & \yellow{00}0 & \yellow{00}0 & \yellow{00}0 & \yellow{00}0 & \yellow{00}1 & \yellow{00}0 & \yellow{00}0 & \yellow{00}0 & \yellow{00}1 & \yellow{00}3 & \yellow{00}0 & \yellow{00}0 & \yellow{00}0 & \yellow{00}0 & \yellow{00}0 & \yellow{00}0 \\
& ccomp & \yellow{00}5 & \yellow{00}7 & \yellow{00}0 & \yellow{30}4 & \yellow{00}7 & \yellow{00}10 & \yellow{00}8 & \yellow{00}2 & \yellow{00}2 & \yellow{00}2 & \yellow{00}9 & \yellow{00}0 & \yellow{00}0 & \yellow{00}0 & \yellow{00}0 & \yellow{00}0 & \yellow{00}0 \\
& xcomp & \yellow{00}6 & \yellow{00}9 & \yellow{00}0 & \yellow{10}3 & \yellow{00}15 & \yellow{00}5 & \yellow{00}5 & \yellow{10}9 & \yellow{00}5 & \yellow{00}11 & \yellow{00}17 & \yellow{00}1 & \yellow{10}16 & \yellow{00}3 & \yellow{00}1 & \yellow{00}2 & \yellow{10}10 \\
& obl   & \yellow{80}88 & \yellow{50}62 & \yellow{100}5 & \yellow{100}14 & \yellow{70}92 & \yellow{65}53 & \yellow{40}25 & \yellow{70}44 & \yellow{70}77 & \yellow{80}202 & \yellow{80}224 & \yellow{60}19 & \yellow{100}121 & \yellow{50}17 & \yellow{70}26 & \yellow{7}6 & \yellow{13}11 \\
& advcl & \yellow{00}10 & \yellow{00}9 & \yellow{90}4 & \yellow{70}8 & \yellow{30}47 & \yellow{15}21 & \yellow{30}19 & \yellow{10}10 & \yellow{10}18 & \yellow{00}14 & \yellow{3}41 & \yellow{00}3 & \yellow{10}15 & \yellow{00}2 & \yellow{10}6 & \yellow{10}7 & \yellow{00}2 \\
& advmod & \yellow{00}0 & \yellow{00}3 & \yellow{00}0 & \yellow{00}0 & \yellow{00}1 & \yellow{00}0 & \yellow{00}0 & \yellow{00}0 & \yellow{00}1 & \yellow{00}0 & \yellow{00}0 & \yellow{00}0 & \yellow{00}1 & \yellow{00}0 & \yellow{00}0 & \yellow{00}0 & \yellow{00}0 \\
& nmod  & \yellow{100}100 & \yellow{100}140 & \yellow{50}2 & \yellow{90}12 & \yellow{100}181 & \yellow{70}57 & \yellow{70}47 & \yellow{100}58 & \yellow{100}148 & \yellow{100}276 & \yellow{100}386 & \yellow{85}29 & \yellow{70}72 & \yellow{100}43 & \yellow{100}35 & \yellow{50}19 & \yellow{50}48 \\
& appos & \yellow{20}26 & \yellow{65}89 & \yellow{00}0 & \yellow{00}2 & \yellow{60}85 & \yellow{100}108 & \yellow{10}11 & \yellow{30}23 & \yellow{30}41 & \yellow{30}72 & \yellow{50}112 & \yellow{20}9 & \yellow{5}12 & \yellow{00}6 & \yellow{10}6 & \yellow{00}1 & \yellow{30}20 \\
& nummod & \yellow{00}1 & \yellow{00}0 & \yellow{00}0 & \yellow{00}0 & \yellow{00}0 & \yellow{00}0 & \yellow{00}0 & \yellow{00}0 & \yellow{00}0 & \yellow{00}0 & \yellow{00}0 & \yellow{00}0 & \yellow{00}0 & \yellow{00}0 & \yellow{00}0 & \yellow{00}0 & \yellow{00}0 \\
& acl   & \yellow{50}40 & \yellow{10}24 & \yellow{00}0 & \yellow{00}0 & \yellow{10}39 & \yellow{30}30 & \yellow{10}10 & \yellow{40}25 & \yellow{40}48 & \yellow{5}33 & \yellow{10}74 & \yellow{00}0 & \yellow{5}11 & \yellow{70}24 & \yellow{00}2 & \yellow{30}13 & \yellow{20}15 \\
& amod  & \yellow{00}5 & \yellow{00}1 & \yellow{00}0 & \yellow{00}2 & \yellow{8}31 & \yellow{00}5 & \yellow{00}3 & \yellow{00}3 & \yellow{00}5 & \yellow{00}2 & \yellow{00}3 & \yellow{00}0 & \yellow{00}3 & \yellow{00}2 & \yellow{00}0 & \yellow{00}3 & \yellow{00}4 \\
& det   & \yellow{00}0 & \yellow{00}0 & \yellow{00}0 & \yellow{00}0 & \yellow{00}0 & \yellow{00}0 & \yellow{00}0 & \yellow{00}0 & \yellow{00}0 & \yellow{00}0 & \yellow{00}1 & \yellow{00}0 & \yellow{00}0 & \yellow{00}0 & \yellow{00}0 & \yellow{00}0 & \yellow{00}0 \\
& conj  & \yellow{00}1 & \yellow{00}4 & \yellow{00}0 & \yellow{00}0 & \yellow{00}3 & \yellow{00}1 & \yellow{00}0 & \yellow{00}1 & \yellow{00}2 & \yellow{00}3 & \yellow{00}11 & \yellow{00}0 & \yellow{00}1 & \yellow{00}1 & \yellow{00}0 & \yellow{00}0 & \yellow{00}0 \\
& flat  & \yellow{00}2 & \yellow{00}3 & \yellow{00}0 & \yellow{00}0 & \yellow{00}1 & \yellow{00}12 & \yellow{00}8 & \yellow{00}0 & \yellow{00}2 & \yellow{00}11 & \yellow{00}37 & \yellow{10}8 & \yellow{00}7 & \yellow{00}1 & \yellow{00}0 & \yellow{00}0 & \yellow{00}3 \\
& compound & \yellow{30}29 & \yellow{10}24 & \yellow{00}0 & \yellow{40}5 & \yellow{45}70 & \yellow{28}27 & \yellow{00}5 & \yellow{00}7 & \yellow{50}54 & \yellow{10}53 & \yellow{5}57 & \yellow{00}2 & \yellow{00}9 & \yellow{00}2 & \yellow{00}5 & \yellow{10}10 & \yellow{30}22 \\
& list  & \yellow{00}0 & \yellow{00}1 & \yellow{00}0 & \yellow{00}0 & \yellow{00}2 & \yellow{00}2 & \yellow{00}0 & \yellow{00}0 & \yellow{00}0 & \yellow{00}0 & \yellow{00}2 & \yellow{00}0 & \yellow{00}2 & \yellow{00}0 & \yellow{00}0 & \yellow{00}0 & \yellow{00}0 \\
& parataxis & \yellow{00}5 & \yellow{00}7 & \yellow{00}0 & \yellow{00}0 & \yellow{8}30 & \yellow{8}14 & \yellow{00}0 & \yellow{00}0 & \yellow{00}14 & \yellow{00}5 & \yellow{00}17 & \yellow{00}1 & \yellow{00}8 & \yellow{00}1 & \yellow{00}5 & \yellow{00}0 & \yellow{00}1 \\
& orphan & \yellow{00}1 & \yellow{00}0 & \yellow{00}0 & \yellow{00}0 & \yellow{00}0 & \yellow{00}0 & \yellow{00}0 & \yellow{00}0 & \yellow{00}0 & \yellow{00}0 & \yellow{00}0 & \yellow{00}0 & \yellow{00}0 & \yellow{00}0 & \yellow{00}0 & \yellow{00}0 & \yellow{00}0 \\
& punct & \yellow{00}0 & \yellow{00}1 & \yellow{00}0 & \yellow{00}0 & \yellow{00}0 & \yellow{00}0 & \yellow{00}0 & \yellow{00}0 & \yellow{00}0 & \yellow{00}0 & \yellow{00}0 & \yellow{00}0 & \yellow{00}0 & \yellow{00}0 & \yellow{00}0 & \yellow{00}0 & \yellow{00}0 \\

\bottomrule
\end{tabularx}
\caption{\textbf{UD Label Distribution Over the Shortest Dependency Paths per Relation Type.} Statistics of the UD labels which are on the shortest dependency path between two entities divided by the 17 relation types of CrossRE~\cite{bassignana-plank-2022-crossre}.}
\label{fig:per-label-ud}
\end{table*}

\begin{table*}[]
    \centering
    \setlength{\tabcolsep}{0.1em}
    \tiny
    \begin{tabularx}{\textwidth}{|c|c|YYYYYYYYYYYYYYYYY|}
\toprule
\multicolumn{2}{c|}{} & related-to & artifact & cause-eff & compare & gen-aff & named & opposite & origin & part-of & physical & role & social & temporal & topic & type-of & usage & win-def \\
\midrule
\multirow{11}{*}{\rotatebox{90}{\textbf{\fontsize{5}{5}\selectfont Shortest Dependency Path Length  \vspace{5pt}}}}
& 1 & 3 & \green{50}43 & 0 & 0 & \green{90}73 & \green{90}92 & 1 & \green{70}19 & \green{30}26 & \green{60}72 & \green{30}64 & 0 & \green{10}5 & \green{50}7 & \green{90}17 & 0 & \green{30}12 \\
& 2 & \green{50}29 & \green{50}43 & 0 & \green{50}3 & \green{70}59 & \green{10}23 & \green{90}25 & \green{80}28 & \green{70}51 & \green{90}85 & \green{70}127 & \green{70}10 & \green{90}41 & \green{30}6 & \green{50}11 & \green{50}6 & \green{70}31 \\
& 3 & \green{70}36 & \green{90}71 & \green{90}2 & \green{10}1 & \green{30}33 & 14 & \green{70}21 & \green{90}29 & \green{90}57 & \green{70}75 & \green{90}136 & \green{90}17 & \green{70}30 & \green{90}14 & 3 & \green{90}14 & \green{90}34 \\
& 4 & \green{90}42 & \green{30}24 & \green{90}2 & \green{70}4 & \green{50}52 & \green{10}20 & \green{50}13 & \green{30}12 & \green{50}37 & \green{50}41 & \green{50}104 & \green{50}7 & \green{60}28 & \green{30}6 & \green{90}17 & \green{70}12 & \green{50}17 \\
& 5 & \green{30}17 & \green{10}12 & 0 & \green{90}5 & \green{30}36 & 12 & \green{50}14 & \green{10}8 & \green{10}18 & \green{30}28 & \green{10}33 & \green{30}5 & \green{40}10 & \green{70}8 & 2 & \green{30}4 & 3 \\
& 6 & \green{10}9 & 7 & 0 & \green{70}4 & \green{10}18 & 6 & \green{10}4 & 5 & 8 & \green{10}12 & 10 & 0 & 2 & 1 & 2 & 1 & 2 \\
& 7 & 4 & 0 & 0 & 0 & 4 & 3 & 0 & 0 & 2 & 6 & 4 & 0 & \green{10}5 & 0 & 0 & 0 & \green{10}5 \\
& 8 & 0 & 0 & 0 & 0 & 0 & 3 & 1 & 0 & 0 & 1 & 1 & 0 & 3 & 0 & 0 & 0 & 0 \\
& 9 & 0 & 1 & 0 & 0 & 0 & 1 & 0 & 0 & 0 & 0 & 2 & 0 & 0 & 0 & 0 & 0 & 0 \\
& 10 & 0 & 0 & 0 & 0 & 0 & 2 & 0 & 0 & 0 & 0 & 0 & 0 & 0 & 0 & 0 & 0 & 0 \\
& 11 & 0 & 0 & 0 & 0 & 0 & 2 & 0 & 0 & 0 & 0 & 0 & 0 & 0 & 0 & 0 & 0 & 0 \\
\bottomrule
    \end{tabularx}
    \caption{\textbf{Shortest Dependency Path Length per Relation Type.} Statistics of the shortest dependency path length between two semantic entities divided by the 17 relation types of CrossRE~\cite{bassignana-plank-2022-crossre}.}
    \label{fig:per-label-sdp}
\end{table*}

\section{Reproducibility}
\label{app:reproducibility}

We report in Table~\ref{tab:reproducibility} the hyperparameter setting of our RE model (see Section~\ref{sec:exp-setup}).
All experiments were ran on an NVIDIA\textsuperscript{®} A100 SXM4 40 GB GPU and an AMD EPYC\textsuperscript{™} 7662 64-Core CPU.
Within this computation infrastructure the baseline converges in $\sim7$ minutes. The the \textit{syntax pre-training} step takes $\sim10$ minutes, to which we have to add $\sim7$ minutes in order to obtain the complete training time.

\begin{table}
    \centering
    \resizebox{0.9\columnwidth}{!}{
    \begin{tabular}{r|l}
        \toprule
        \textbf{Parameter} & \textbf{Value} \\
        \midrule
        Encoder & \texttt{bert-base-cased} \\
        Classifier & 1-layer FFNN \\
        Loss & Cross Entropy\\
        Optimizer & Adam optimizer \\
        Batch size & 12, 24  \\
        Learning rate & $1e^{-5}$ (pre-train) \\
        Learning rate & $2e^{-5}$ (fine-tuning) \\
        Seeds & 4012, 5096, 8878, 8857, 9908 \\
        \bottomrule
    \end{tabular}}
    \caption{\textbf{Hyperparameters Setting.} Model details for reproducibility of the baseline.}
    \label{tab:reproducibility}
\end{table}

We train MaChAmp v0.4 on the English Web Treebank v2.10 with XLM-R large~\cite{conneau-etal-2020-unsupervised} as language model with all default hyperparameters of MaChAmp.

\section{Handling of \texttt{Conj}}
\label{app:conj}

\begin{figure*}
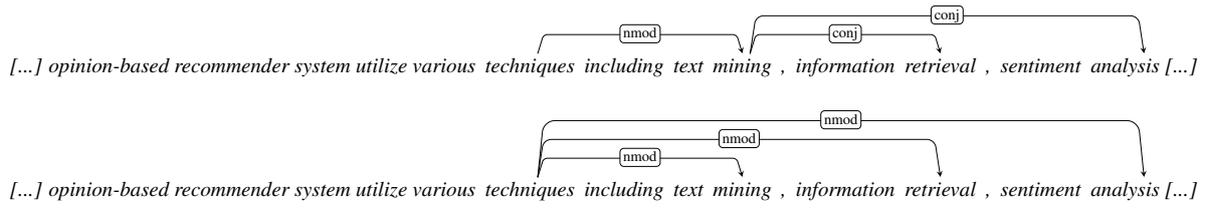

\resizebox{\textwidth}{!}{
\begin{dependency}
\begin{deptext}
\textit{[...] opinion-based recommender system utilize various} \& \textit{techniques} \& \textit{including} \& \textit{text} \& \textit{mining} \& \textit{,} \& \textit{information} \& \textit{retrieval} \& \textit{,} \& \textit{sentiment} \& \textit{analysis [...]} \\ \\
\end{deptext}
\depedge[edge height=2ex]{2}{5}{nmod}
\depedge[edge height=2ex]{5}{8}{conj}
\depedge[edge height=4ex]{5}{11}{conj}
\end{dependency}}
\\
\resizebox{\textwidth}{!}{
\begin{dependency}
\begin{deptext}
\textit{[...] opinion-based recommender system utilize various} \& \textit{techniques} \& \textit{including} \& \textit{text} \& \textit{mining} \& \textit{,} \& \textit{information} \& \textit{retrieval} \& \textit{,} \& \textit{sentiment} \& \textit{analysis [...]} \\ \\
\end{deptext}
\depedge[edge height=2ex]{2}{5}{nmod}
\depedge[edge height=4ex]{2}{8}{nmod}
\depedge[edge height=6ex]{2}{11}{nmod}
\end{dependency}}
\caption{\label{fig:fixed-conj}\textbf{Example of \texttt{conj} Alteration}. Original UD dependencies (above), and disentangled conjunction dependencies reflecting the semantic relation annotations (below).}
\end{figure*}

In UD, the first element in a conjuncted list governs all other elements of the list via a \texttt{conj} dependency and represents the list syntactically w.r.t.\ the remainder of the sentence.  CrossRE~\cite{bassignana-plank-2022-crossre} relations, on the other hand, directly link the two entities involved in the semantic structure. To account for this difference, we propagate the conjunction dependencies in order to reflect the semantic relations, as shown in Figure~\ref{fig:fixed-conj}.

\section{CrossRE Size}
\label{app:crossre}

We report in Table~\ref{tab:dataset-statistics} the dataset statistics of CrossRE~\cite{bassignana-plank-2022-crossre} including the number of sentences and of relations.

\begin{table}
    \centering
    \resizebox{\columnwidth}{!}{
    \begin{tabular}{r|ccc|c|ccc|c}
    \toprule
    & \multicolumn{4}{c|}{\textsc{sentences}} & \multicolumn{4}{c}{\textsc{relations}} \\
    \midrule
    & train & dev & test & \textbf{tot.} & train & dev & test & \textbf{tot.} \\
    \midrule
    news & 164 & 350 & 400 & 914 & 175 & 300 & 396 & 871 \\
    politics & 101 & 350 & 400 & 851 & 502 & 1,616 & 1,831 & 3,949 \\
    science & 103 & 351 & 400 & 854 & 355 & 1,340 & 1,393 & 3,088 \\
    music & 100 & 350 & 399 & 849 & 496 & 1,861 & 2,333 & 4,690 \\
    literature & 100 & 400 & 416 & 916 & 397 & 1,539 & 1,591 & 3,527 \\
    AI & 100 & 350 & 431 & 881 & 350 & 1,006 & 1,127 & 2,483 \\
    \midrule
    \textbf{tot.} & 668 & 2,151 & 2,446 & \textbf{5,265} & 2,275 & 7,662 & 8,671 & \textbf{18,608} \\
    \bottomrule
    \end{tabular}}
    \caption{\textbf{CrossRE Statistics.} Number of sentences and number of relations for each domain of CrossRE~\cite{bassignana-plank-2022-crossre}.
    }
    \label{tab:dataset-statistics}
\end{table}

\section{Syntax Pre-training Performance}
\label{app:paser}

\begin{figure}
    \centering
    \includegraphics[width=\columnwidth]{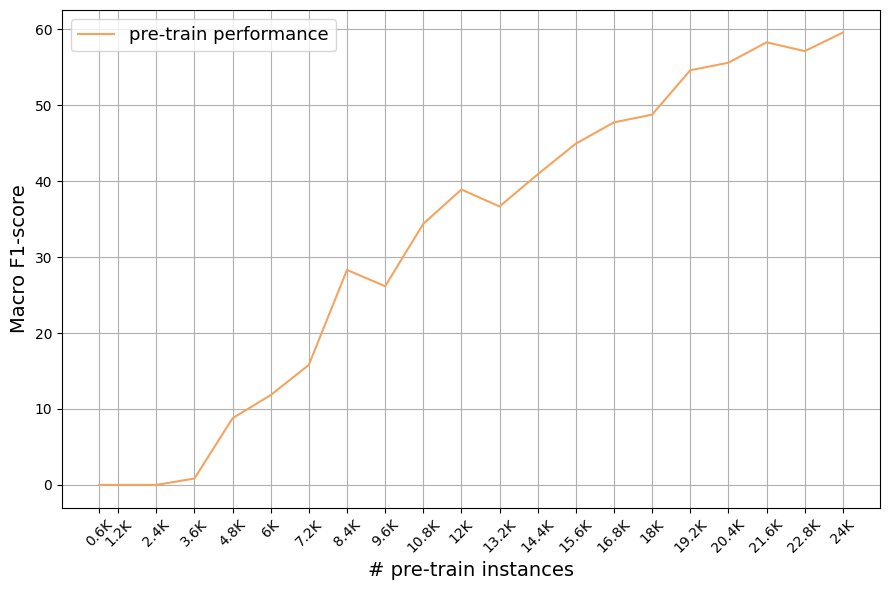}
    \caption{\textbf{Pre-train Performance.} Pre-train performance of the RE model over increasing amounts of dependency instances}
    \label{fig:parser}
\end{figure}

Figure~\ref{fig:parser} reports the performance of the RE model during the \textit{syntax pre-training} phase, over increasing amounts of pre-training dependency instances. The scores are computed on a set including 600 sentences (100 per domain) not overlapping with the train set used in the syntax pre-training phase.

\end{document}